\pdfoutput=1

\documentclass[11pt]{article}

\usepackage[preprint]{acl}

\usepackage{times}
\usepackage{latexsym}

\usepackage[T1]{fontenc}

\usepackage[utf8]{inputenc}

\usepackage{microtype}

\usepackage{inconsolata}

\usepackage{booktabs}
\usepackage{multirow}
\usepackage{xfrac}
\usepackage{float}

\renewcommand\cite{\citep}
\usepackage{caption}

\usepackage{mdframed}
\usepackage{listings}
\usepackage{caption}
\newcommand{\distance}{7pt}
\usepackage{tabularray}
\UseTblrLibrary{booktabs}
\SetTblrInner[booktabs]{abovesep=0pt, belowsep=0pt, rowsep=0.2pt}
\SetTblrInner[booktabs]{cells = {cmd=\small}}

\NewTableCommand\seprule{\specialrule{\lightrulewidth,gray8}{2pt}{2pt}}
\NewTableCommand\uniquerule{\specialrule{\lightrulewidth,gray7,dashed}{2pt}{2pt}}
\definecolor{lightb}{RGB}{235,245,255}

\usepackage{graphicx}

\usepackage{pifont}%

\usepackage[nointegrals]{wasysym}
\usepackage{enumitem} %
\setlist[itemize]{leftmargin=*}

\usepackage{xspace}
\usepackage{xcolor}
\usepackage{mathtools}

\usepackage[capitalise, noabbrev]{cleveref}
\crefformat{section}{\S#2#1#3}
\Crefformat{section}{\S#2#1#3}

\usepackage{fontawesome}

\newcommand\COMMENT[1]{}

\usepackage[commandnameprefix=always]{changes}
\newcommand{\revisionComment}[1]{\textcolor{blue}{\textit{[#1]}}}
\setanonymousname{Yuxiang}
\setauthormarkuptext{name}
\setcommentmarkup{\revisionComment{Yuxiang: #1}}

\usepackage[normalem]{ulem}
\definecolor{applegreen}{rgb}{0.55, 0.71, 0.0}

\newcommand\passat[1]{\mbox{pass@{#1}}}
\newcommand\Passat[1]{\mbox{Pass@{#1}}}

\newcommand\ossinstruct{\textsc{OSS-Instruct}\xspace}
\newcommand\ours{${\mathcal X}$FT\xspace}
\newcommand\oursmoe{MoE$_{\small \textsc{DS}}$\xspace}
\newcommand\oursmerge{${\mathcal X}$FT$_{\small \textsc{DS}}$\xspace}
\newcommand\oursscalemoe{MoE$_{\small \textsc{DS-6.7B}}$\xspace}
\newcommand\oursscalemerge{${\mathcal X}$FT$_{\small \textsc{DS-6.7B}}$\xspace}
\newcommand\stablemoe{MoE$_{\small \textsc{STABLE}}$\xspace}
\newcommand\stablemerge{${\mathcal X}$FT$_{\small \textsc{STABLE}}$\xspace}
\newcommand\baselineds{SFT$_{\small \textsc{DS}}$\xspace}
\newcommand\baselinedsscale{SFT$_{\small \textsc{DS-6.7B}}$\xspace}
\newcommand\baselinestable{SFT$_{\small \textsc{STABLE}}$\xspace}
\newcommand\baselinetinyllama{SFT$_{\small \textsc{TL}}$\xspace}
\newcommand\tinyllamamoe{MoE$_{\small \textsc{TL}}$\xspace}
\newcommand\tinyllamamerge{${\mathcal X}$FT$_{\small \textsc{TL}}$\xspace}
\newcommand\ewads{EWA$_{\small \textsc{DS}}$\xspace}

\newcommand\evolinstruct{Evol-Instruct\xspace}
\newcommand\evolcode{\texttt{evol-codealpaca-v1}\xspace}
\newcommand\evolgeneral{\texttt{evol-instruct-70k}\xspace}
\newcommand\selfinstruct{\textsc{Self-Instruct}\xspace}

\newcommand\llm{LLM\xspace}
\newcommand\llmfull{Large Language Model\xspace}
\newcommand\moe{MoE\xspace}
\newcommand\moefull{Mixture-of-Experts\xspace}
\newcommand\ewafull{Experts Weights Averaging\xspace}
\newcommand\sparseupcycle{sparse upcycling\xspace}
\newcommand\deepseekmoe{DeepSeekMoE\xspace}
\newcommand\mocle{MoCLE\xspace}
\newcommand\loramoe{LoRAMoE\xspace}
\newcommand\pesc{PESC\xspace}
\newcommand\lora{LoRA\xspace}
\newcommand\modelsoup{Model Soups\xspace}
\newcommand\ewa{EWA\xspace}

\newcommand\bigcodeharness{\texttt{bigcode-evaluation-harness}}

\newcommand\nsamples{\texttt{num\_samples}}
\newcommand\temperature{\texttt{temperature}}
\newcommand\topp{\texttt{top\_p}}
\newcommand\maxLen{\texttt{max\_length}}

\newcommand{\ie}{\emph{i.e.,}\xspace}

\newcommand\chatgpt{ChatGPT\xspace}
\newcommand\gptthreefive{GPT-3.5\xspace}

\newcommand\dscoderbase{DeepSeek-Coder-Base\xspace}
\newcommand\dscoderinst{DeepSeek-Coder-Instruct\xspace}
\newcommand\stablecoder{\textsc{stable-code}\xspace}
\newcommand\tinyllama{TinyLlama\xspace}

\newcommand\humaneval{HumanEval\xspace}
\newcommand\humanevalp{HumanEval+\xspace}
\newcommand\evalplus{EvalPlus\xspace}
\newcommand\mbpp{MBPP\xspace}
\newcommand\mbppp{MBPP+\xspace}

\newcommand\dsonek{DS-1000\xspace}

\newcommand\multiple{MultiPL-E\xspace}

\title{
\ours: Unlocking the Power of Code Instruction Tuning \\
by Simply Merging Upcycled Mixture-of-Experts}

\author{Yifeng Ding, Jiawei Liu, Yuxiang Wei, Terry Yue Zhuo, Lingming Zhang \\
  University of Illinois Urbana-Champaign \\
  \texttt{\{yifeng6, lingming\}@illinois.edu} \\}

\begin{document}
\maketitle

\begin{abstract}
We introduce \textbf{\ours}, a simple yet powerful training scheme, by simply merging upcycled \moefull (\moe) to unleash the performance limit of instruction-tuned code \llmfull{s} (\llm{s}).
While vanilla \sparseupcycle fails to improve instruction tuning, \ours introduces a shared expert mechanism with a novel routing weight normalization strategy into \sparseupcycle, 
which significantly boosts instruction tuning.
After fine-tuning the upcycled \moe model, 
\ours introduces a learnable model merging mechanism to compile the upcycled \moe model back to a dense model,
achieving upcycled \moe{-level} performance with only dense-model compute.
By applying \ours to a 1.3B model, 
we create a new state-of-the-art tiny code \llm{} (<3B) with 67.1 and 64.6 \passat{1} on \humaneval and \humanevalp{} respectively.
With the same data and model architecture,
\ours improves supervised fine-tuning (SFT) by 13\% on \humanevalp, along with consistent improvements from 2\% to 13\% on \mbppp, \multiple, and \dsonek, demonstrating its generalizability.
\ours is fully orthogonal to existing techniques such as \evolinstruct{} and \ossinstruct{}, opening a new dimension for improving code instruction tuning. Codes are available at \url{https://github.com/ise-uiuc/xft}.

\end{abstract}

\section{Introduction}
Program synthesis (or code generation) is a long-standing problem explored since the early days of computer science~\cite{manna1971toward}. 
Recently, instruction tuning of code \llmfull{s} (\llm{s}) has been used to improve many coding tasks~\cite{codealpaca, luo2023wizardcoder, wei2023magicoder}, such as text-to-code generation~\cite{chen2021evaluating, austin2021program}, code completion~\cite{cassano2022multiple}, and data science engineering~\cite{lai2022ds1000}. 

\begin{figure}[t]
\centering
\includegraphics[width=1.0\linewidth]{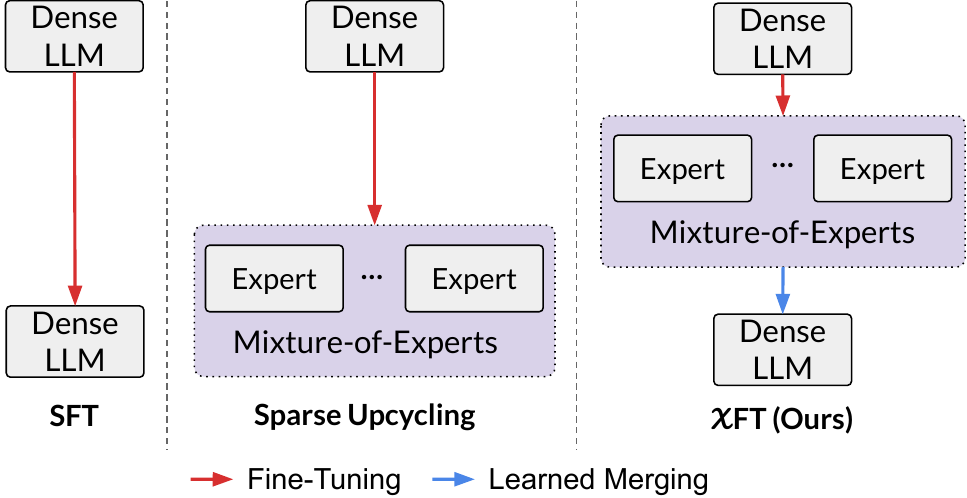}
\caption{Overview of SFT, \sparseupcycle, and \ours.}
\label{fig:comparison}
\end{figure}

A typical instruction tuning flow involves two steps~\cite{zhang2023instruction}: 
(i) curating an instruction dataset of instruction-output pairs, where the instruction reflects human intents in natural language and the output includes target code snippets that correspond to the intent; 
and 
(ii) supervised fine-tuning of pre-trained \llm on the instruction dataset. 
In the realm of code instruction tuning, most recent works have been focusing on curating high-quality instruction datasets.
For example, \textit{Code \evolinstruct}~\cite{luo2023wizardcoder} uses \chatgpt to obtain complex synthetic code instructions with heuristic prompts, while \ossinstruct~\cite{wei2023magicoder} prompts \chatgpt to generate new coding problems by drawing inspiration from open source code snippets.
Since existing works focus on the data perspectives of instruction tuning, 
they all follow the standard SFT, leaving room for exploring advanced training schemes.

We argue that prior works largely overlook the possibility of improving code instruction tuning by advancing existing training schemes.
Figure~\ref{fig:comparison} depicts supervised fine-tuning (SFT), which directly uses the pre-trained weights and architecture for fine-tuning.
The model is \emph{dense} here because all parameters are activated to predict the next token (assuming it is a decoder-only \llm{}).
In contrast to fine-tuning a \emph{dense} model, 
following the scaling laws~\cite{kaplan2020scaling} (\ie more parameters, better performance),
\sparseupcycle~\cite{komatsuzaki2023sparse} is proposed to efficiently upgrade the model size by upcycling a dense \llm to a sparsely activated \moefull (\moe) model.
An \moe model is efficient because its prediction of the next token only invokes a subset of parameters (\ie experts) and thus is \emph{sparsely activated}.
For example, Mixtral-8x7B~\cite{jiang2024mixtral}, compared to a dense 7B model, 
uses approximately $8\times$ parameters and $2\times$ computation, \ie only 2 out of 8 experts are dynamically selected to compute the next token.
However, there are two key limitations when using \sparseupcycle in instruction tuning:
(i) \emph{Slow scaling:} it is reported that \sparseupcycle improves the performance of dense models marginally with limited training steps, requiring orders of magnitude of extra compute to achieve decent improvement~\cite{komatsuzaki2023sparse}; (ii) \emph{Inference cost:}
although \moe is more efficient than directly scaling up the size of dense \llm{s},
\moe is still expensive, especially at inference, 
as it introduces significantly more parameters (\ie memory) and computes during inference, compared to its dense counterparts.

In this paper, we propose \textbf{\ours}: 
by simply merging upcycled \moe models, we push the performance limit of instruction-tuned code \llm{s}.
While vanilla \sparseupcycle fails to improve instruction tuning efficiently~\cite{komatsuzaki2023sparse},
\ours addresses this challenge by isolating one expert as the shared expert among all the other experts in each \moe layer, inspired by \deepseekmoe~\cite{dai2024deepseekmoe} and \mocle~\cite{gou2024mixture}.
\ours also proposes a novel routing weight normalization strategy to eliminate scale mismatch between the upcycled \moe layer with the shared expert and the original dense layer, which will otherwise lead to performance degradation~\cite{wu2022residual}. 
After the upcycled \moe model finishes its SFT phase, motivated by \modelsoup~\cite{wortsman2022model}, 
\ours uses a learnable model merging mechanism to output a dense model by merging all the expert networks in the upcycled \moe, 
\ie the final dense model is of the same model structure and size as the original pre-trained model,
achieving similar performance without paying extra inference cost as the \sparseupcycle.
With only 1.3B parameters, \ours achieves 67.1 \passat{1} on \humaneval and 64.6 \passat{1} on \humanevalp, which is the new state-of-the-art for tiny code \llm{s} (<3B).
Compared with SFT, \ours achieves 13\% improvement on \humanevalp.
Surprisingly, our learnable merging mechanism can preserve or even further boost the performance of the upcycled \moe with only around $\sfrac{1}{8}\times$ parameters! 
We conclude our contribution as follows:
\begin{itemize}[leftmargin=1em]
    \setlength{\parskip}{2pt}
    \setlength\itemsep{0pt}
    \item \textbf{Dimension:} 
    We open a new dimension of improving instruction tuning of code \llm{s} by advancing its training scheme, using enhanced \sparseupcycle and learnable model merging mechanism, which neither changes the final model structure nor requires more training data.
    \item \textbf{Technique:} 
    We present \ours, a new training scheme for code instruction tuning. \ours involves two steps: \emph{upcycling} and \emph{merging}. A pre-trained dense \llm is first upcycled into an \moe with the shared expert setting and then fine-tuned on the instruction dataset. We propose a novel routing weight normalization strategy to avoid the performance degradation caused by the scale mismatch problem.
    In addition, we introduce the first learnable mechanism 
    for merging the upcycled \moe into a dense model, eliminating additional inference overhead while preserving or even improving the upcycled \moe performance.
    \item \textbf{Results:} With only 1.3B parameters, \ours achieves 67.1 \passat{1} on \humaneval and 64.6 \passat{1} on \humanevalp, which is the new state-of-the-art for tiny code \llm{s} (<3B). Compared with normal SFT, \ours achieves a significant 13\% improvement on \humanevalp! \ours also achieves consistent improvements from 2\% to 13\% on \mbpp, \multiple, and \dsonek over SFT, demonstrating its generalizability.
\end{itemize}

\section{Related Work}
\subsection{\moefull}
\moefull (\moe) can efficiently scale up model sizes with only sub-linear increases in computation~\cite{shazeer2017outrageously}. 
Compared with the standard Transformer, \moe replaces each Feed-Forward Network (FFN) layer with an \moe layer, which uses $N$ (\ie multiple) expert networks that are structurally equivalent to the original FFN layer and uses a router that directs each input token to $K$ out of $N$ expert networks. 
Formally, for the $l$-th \moe layer, output hidden state $\mbox{\textbf{h}}_t^l$ of the $t$-th input token is computed as follows~\cite{dai2024deepseekmoe}:
\begin{equation}\label{formula:moe}
\begin{split}
\mbox{\textbf{h}}_t^l &= \sum_{i=1}^{N}(g_{i,t}\mbox{FFN}_i(\mbox{\textbf{u}}_t^l)) + \mbox{\textbf{u}}_t^l \\
g_{i,t} &= 
\begin{cases}
 s_{i,t} & s_{i,t}\in  \mbox{Topk}(s_t, K) \\
    0 & \text{otherwise}
\end{cases} \\
s_t &= \{s_{i,t} \mid 1\leq i\leq N\} \\
s_{i,t} &= \mbox{Softmax}_i({\mbox{\textbf{u}}_t^l}^T\mbox{\textbf{e}}_i^l)
\end{split}
\end{equation}
where $g_{i,t}$ refers to the gate value for the $i$-th expert given the $t$-th token, $\mbox{FFN}_i(\cdot)$ refers to the $i$-th expert, $\mbox{\textbf{u}}_t^l$ refers to the hidden states of the $t$-th token which is the input of the $l$-th \moe layer, $s_{i,t}$ refers to the affinity score between the $i$-th expert and the $t$-th token, $\mbox{Topk}(S, K)$ refers to a function extracting $K$ highest scores out of $S$, and $\mbox{\textbf{e}}_i^l$ refers to the centroid of the $i$-th expert in the $l$-th \moe layer. By definition, each token will only be computed in the top $K$ experts among all the $N$ experts and such sparsity assures the efficiency of \moe.

Recently, many works have been proposed to scale model sizes with \moe architecture~\cite{lepikhin2020gshard, du2022glam, fedus2022switch, jiang2024mixtral, xue2024openmoe}. While 
most \moe models are trained from scratch, \sparseupcycle~\cite{komatsuzaki2023sparse} is proposed to initialize \moe models based on pre-trained dense models, which can efficiently reduce the computational costs of training \moe models, compared with training \moe models from scratch. 
Specifically, \sparseupcycle constructs a new \moe model by initializing each expert of each \moe layer as a copy of the original FFN layer in the dense model, while directly copying the remaining layers from the dense model to the new \moe model.

\subsection{Instruction Tuning}
Instruction tuning is designed to improve the instruction-following ability of \llm{s} by fine-tuning them on the instruction datasets in a supervised fashion~\cite{wei2022finetuned}. 
The quality of the instruction dataset is significant for the effectiveness of instruction tuning and researchers have proposed multiple methods to improve data quality. For example, \textsc{\selfinstruct}~\cite{wang2023selfinstruct} synthesizes high-quality instruction data by prompting a foundation \llm with carefully designed prompts. To improve \textsc{\selfinstruct}, \evolinstruct~\cite{xu2023wizardlm} enhances the complexity and diversity of the instruction dataset by prompting \chatgpt with heuristic prompts. \textsc{\ossinstruct}~\cite{wei2023magicoder} queries \chatgpt to generate instruction-output pairs by getting inspiration from real-world code snippets.

Recently, some parameter-efficient fine-tuning techniques have been proposed to use \moe for better instruction tuning. For example, \loramoe~\cite{dou2023loramoe} and \mocle~\cite{gou2024mixture} propose \moe-like modules that are constructed with Low-Rank Adaptations (\lora) to improve instruction tuning, while \pesc~\cite{wu2024parameterefficient} proposes to integrate adapters into \moe that are upcycled from dense models. Unlike these works, \ours focuses on full-parameter fine-tuning, which is proven generally stronger than parameter-efficient fine-tuning~\cite{chen2022revisiting}.

\subsection{Weight Averaging}\label{sec:weight_averaging}
Weight averaging is a commonly used technique to improve the performance of deep learning models. For example, \modelsoup~\cite{wortsman2022model} averages the weights of multiple models that are initialized from the same pre-trained model but finetuned with different hyperparameter configurations to improve the accuracy and robustness of the fine-tuned model. However, only a few works have been proposed to merge experts of an \moe layer to a normal FFN layer with weight averaging to reduce both parameter and computation overhead of inference. For example, OneS~\cite{xue2022student} proposes several simple weight averaging methods to merge expert networks of a BERT-based \moe model. Closely related to our work, \ewafull (\ewa)~\cite{huang2023experts} proposes to convert an \moe model to a dense model with two steps: (i) During \moe training, \ewa conducts weighted averaging of all the expert weights after each weight update of \moe, which is based on a manually-crafted hyperparameter $\beta$; (ii) After training, \ewa converts each \moe layer into an FFN layer by uniformly averaging the experts. 

Different from all the aforementioned existing works, \ours is the first work proposing a \textbf{learnable} mechanism to merge expert networks in the upcycled \moe model. Furthermore, while the training scheme of \ewa is deeply coupled to a specific \moe architecture, \ours can be easily adapted to different \moe architectures by only adjusting the final merging process. 
In addition, unlike \ewa, \ours does not introduce any hyperparameters into the training of large \moe models, 
significantly reducing the computational resources for hyperparameter searching. 
Our empirical results in Section \ref{sec:experiment} also demonstrate the clear advantage of \ours.

\section{\ours}
\begin{figure*}[t]
\centering
\includegraphics[width=0.8\linewidth]{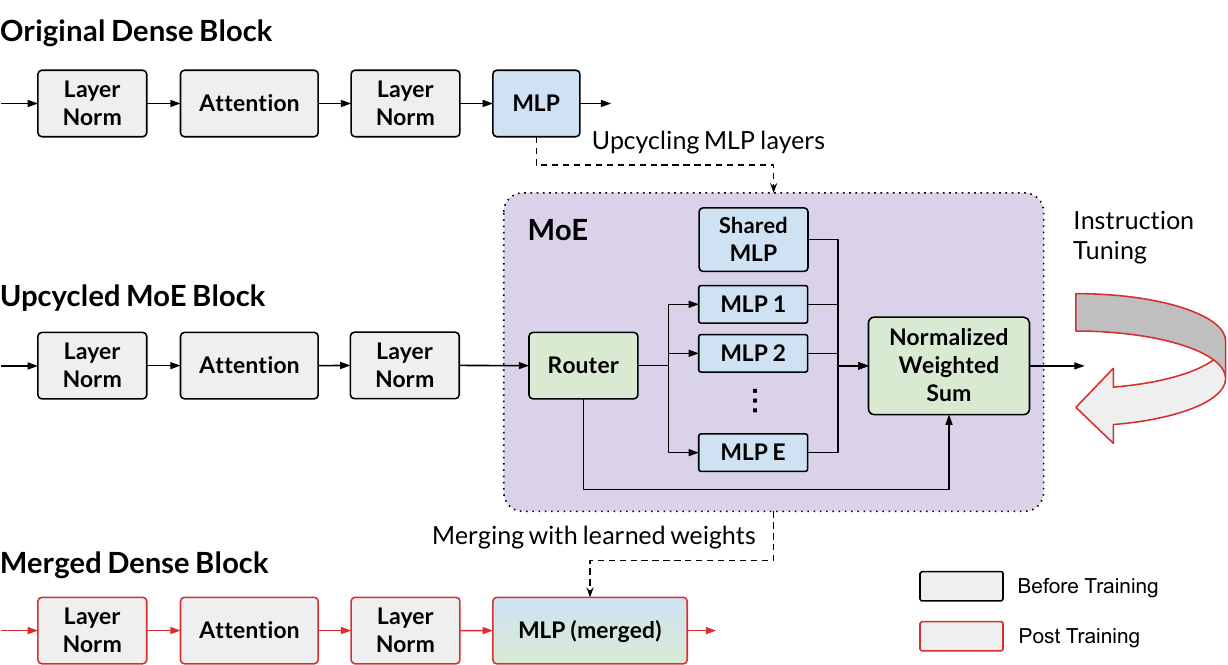}
\caption{Overview of \ours.
}
\label{fig:overview}
\end{figure*}

We describe the details of \ours in this section. There are two steps in our framework: \textit{upcycling} (Section~\ref{sec:upcycling}) and \textit{merging} (Section~\ref{sec:merging}). During upcycling, we construct an \moefull (\moe) model from the pre-trained dense model, namely \textbf{\oursmoe}, which is then fine-tuned on coding instruction data. For merging, we propose a learnable model merging method to convert the instruction-tuned \oursmoe back to a normal dense model by merging each \moe layer into an FFN layer through weight averaging while directly copying other remaining layers. Consequently, we can obtain \textbf{\oursmerge} that has the same model architecture and size as the original pre-trained dense model, which eliminates all the additional inference overhead brought by the original \sparseupcycle, while preserving or even improving the performance of \oursmoe. Our framework is illustrated in Figure \ref{fig:overview}.

\subsection{Upcycling}\label{sec:upcycling}
Inspired by \sparseupcycle~\cite{komatsuzaki2023sparse}, we convert the pre-trained dense \llm to a new \moe by initializing each expert of each \moe layer as a copy of the original FFN layer in the dense model, while directly copying the remaining layers from the dense model to the new \moe model. However, the performance gain brought by \sparseupcycle is negligible with a limited training budget~\cite{komatsuzaki2023sparse} -- which is exactly the situation we are facing during instruction tuning. Intuitively, it is because each expert in the upcycled \moe model is trained on fewer instruction data than the original dense model does because traditional routers used in \sparseupcycle will assign different tokens to different experts and thus reduce the amount of data each expert is trained on~\cite{gou2024mixture}. Consequently, inspired by \deepseekmoe~\cite{dai2024deepseekmoe} and \mocle~\cite{gou2024mixture}, \ours introduces the shared expert setting into \sparseupcycle to tackle this challenge. We further propose a novel routing weight normalization strategy for \ours to avoid the potential performance degradation caused by the scale mismatch problem~\cite{wu2022residual}.

\subsubsection{Shared Expert for Upcycling}
During upcycling, we isolate one shared expert among all the other normal experts in each \moe layer, where the shared expert will be deterministically assigned to handle all the tokens while other normal experts are assigned by the router. By doing so, the upcycled \moe model can achieve a clear performance boost in instruction tuning, where the shared expert learns general knowledge across the whole instruction dataset while other normal experts learn specific knowledge among different instructions assigned by the router. Formally, the output hidden state $\mbox{\textbf{h}}_t^l$ of the $l$-th \moe layer when processing the $t$-th token can be expressed as:
\begin{equation}\label{formula:upcycling}
\begin{split}
\mbox{\textbf{h}}_t^l &= \sum_{i=1}^{N}(g_{i,t}\mbox{FFN}_i(\mbox{\textbf{u}}_t^l)) + \mbox{\textbf{u}}_t^l \\
g_{i,t} &= 
\begin{cases}
 1-{s_{t}}_{\max} & i=1 \\
 \mbox{Softmax}_i(s_{i,t})\cdot {s_{t}}_{\max} & s_{i,t}\in  {S_t}_K \\
    0 & {\text{otherwise}}
\end{cases} \\
{S_t}_K &= \mbox{Topk}(\{s_{i,t} \mid 1\leq i\leq N\}, K-1) \\
{s_{t}}_{\max} &= \max (\{s_{i,t} \mid 1\leq i\leq N\}) \\
s_{i,t} &= \begin{cases}
    -\infty & i = 1 \\
 \mbox{Softmax}_i({\mbox{\textbf{u}}_t^l}^T\mbox{\textbf{e}}_i^l)& i\geq 2
\end{cases} \\
\end{split}
\end{equation}
where $N$ refers to the total number of experts, $K$ refers to the number of activated experts, $g_{i,t}$ refers to the gate value for the $i$-th expert given the $t$-th token, $\mbox{FFN}_i(\cdot)$ refers to the $i$-th expert, $\textbf{u}_t^l$ refers to the output hidden state of the $l$-th attention layer given the $t$-th token (which is also the input of the $l$-th \moe layer), $s_{i, t}$ refers to the affinity score between the $i$-th expert and the $t$-th token, $s_{t_{\max}}$ refers to the maximum affinity score among all the experts besides the shared expert, $\mbox{Topk}(S,K)$ refers to a function extracting $K$ highest scores out of $S$, $S_{tK}$ refers to a set of $K-1$ highest affinity scores among all the experts besides the shared expert, and $\textbf{e}_i^l$ refers to the centroid of the $i$-th expert in the $l$-th \moe layer.

$\mbox{FFN}_1$ is chosen as the shared expert in each \moe layer and each token will be assigned to top $K$ experts including one shared expert and $K-1$ other normal experts. Compared with the original \sparseupcycle, there are two major differences:
\begin{itemize}[leftmargin=1em]
    \setlength{\parskip}{2pt}
    \setlength\itemsep{0pt}
    \item \textbf{Weighted Shared Expert}. Following \mocle~\cite{gou2024mixture}, with the token-to-expert affinity score $s_{i,t}$, we get the maximum affinity score ${s_{t}}_{\max}$ and use its complement $1-{s_{t}}_{\max}$ as the routing weight of the shared expert.
    \item \textbf{Routing Weight Normalization}. Although the shared expert setting is also used in recent works~\cite{dai2024deepseekmoe, gou2024mixture}, we cannot directly follow their routing strategy because they cannot handle a scale mismatch problem that is unique for \sparseupcycle. The scale mismatch problem is that differences between the scale of the output of the upcycled \moe layer and that of the original FFN layer can cause performance degradation~\cite{wu2022residual}. To handle this problem, we need to ensure the sum of $g_{i,t}$ equals 1, so that the output of the \moe layer matches that of the FFN layer in scale. To do so, we normalize the affinity scores of top $K-1$ normal experts with Softmax and scale their sum to ${s_{t}}_{\max}$ to make sure that the sum of the $g_{i,t}$ of top $K$ experts, including one shared expert and $K-1$ normal experts, equals 1.
\end{itemize}

\subsection{Merging}\label{sec:merging}
We propose a learnable model merging method to convert the large MoE model, namely \oursmoe, back to a dense model \oursmerge. By doing so, we expect \oursmerge to keep the boosted performance gained during upcycling while keeping its model size the same as the original dense model size to avoid any additional inference overhead. Inspired by \modelsoup~\cite{wortsman2022model}, we choose to merge \oursmoe by learning the mixing coefficients that can be used to average the parameters of all experts in each \moe layer to obtain a normal FFN layer, while directly copying other remaining layers. 

Formally speaking, given the weights of $N$ experts at the $l$-th layer $W_1^l, W_2^l, \cdots, W_N^l$, the process of merging each \moe layer to an FFN layer can be stated below:
\begin{equation}\label{formula:merging1}
\begin{split}
\overline{W^l} = \sum_{i=1}^{N}\alpha_{i}^lW_{i}^l
\end{split}
\end{equation}
where $\overline{W^l}$ denotes the merged parameter of all $N$ experts and $\alpha_i^l$ denotes the learnable mixing coefficient of expert $W_i^l$. Mixing coefficients $\alpha$ is parameterized as the output of a softmax, ensuring that $\alpha_i^l$ is positive and $\sum_{i=1}^N\alpha_i^l = 1$. Given input $x$, we denote the output of a neural network with parameters $\theta$ as $f(x; \theta)$. For loss $\mathcal{L}$ and instruction dataset $\{(x_i,y_i)\}_{i=1}^m$, such mixing coefficients $\alpha$ of all the $L$ layers can be learned via:
\begin{equation}\label{formula:merging2}
\begin{split}
\arg\min_{\alpha}\sum_{j=1}^{m}\mathcal{L}(f(x_j; \theta_{o}, (\overline{W^l})_{1:L}), y_i)
\end{split}
\end{equation}
where $\theta_{o}$ refers to all the remaining layers of \oursmoe other than \moe layers.

While the learning process defined in Eq. (\ref{formula:merging2}) is the most intuitive way of learning $\alpha$, our preliminary experiment shows that, due to the shared expert setting, it tends to increase the mixing coefficient of the shared expert at each layer as much as possible to decrease the loss. It is not helpful because, although the shared expert has learned general knowledge across the whole instruction dataset and needs a relatively large mixing coefficient, we still need to keep the scale of the mixing coefficient of other normal experts at a certain level to keep the specific knowledge learned by other normal experts in the merged parameter $\overline{W^l}$.

To solve this issue, we introduce a \textit{shared expert rate} $\lambda$ to fix the mixing coefficient of the shared expert and learn the mixing coefficients of the remaining normal experts which sums to $1-\lambda$ in each layer. By doing so, we can easily control the scale of the mixing coefficient of the shared expert, while still being able to learn the optimal layer-wise mixing coefficients of other normal experts. Let's say $W_1^l$ is the shared expert of the $l$-th layer, then Eq. (\ref{formula:merging1}) and Eq. (\ref{formula:merging2}) can be reformulated as follows:
\begin{gather}
\overline{W^l_\lambda} = \lambda W_{1}^l + \sum_{i=2}^{N}\alpha_{i}^lW_{i}^l\label{formula:merging1new}\\
\arg\min_{\alpha}\sum_{j=1}^{m}\mathcal{L}(f(x_j; \theta_o, (\overline{W^l_\lambda})_{1:L}), y_i)\label{formula:merging2new}
\end{gather}

In practice, we uniformly initialize the mixing coefficients $\alpha$ of all the normal experts as $\frac{1-\lambda}{N-1}$, which is then trained on the same instruction dataset used during upcycling.

\section{Main Evaluation}\label{sec:experiment}
\newcommand{\markclosed}{{\color[HTML]{FE0000} Private}}

\begin{table*}[h]
\centering
\begin{tabular}{@{}lccccc@{}}
\toprule
\multirow{2}{*}{Model}  & \multicolumn{1}{c}{\multirow{2}{*}{Size}} & \multirow{2}{*}{\begin{tabular}[c]{@{}c@{}}Instruction\\ Dataset\end{tabular}} & \multirow{2}{*}{\begin{tabular}[c]{@{}c@{}}Dataset\\ Size\end{tabular}} & \multicolumn{2}{c}{Benchmark}                                    \\ \cmidrule(l){5-6} 
                        & \multicolumn{1}{c}{}                      & \multicolumn{1}{c}{}  & \multicolumn{1}{c}{}                                                                               & \multicolumn{1}{c}{HumanEval (+)} & \multicolumn{1}{c}{MBPP (+)} \\ \midrule
\gptthreefive (May 2023) & -         & {\markclosed} & -         & 73.2 (66.5)    & -            \\ \midrule
\stablecoder             & 3B        & -             & -         & 28.7 (25.6)    & 53.6 (44.1)  \\
\dscoderbase             & 1.3B      & -             & -         & 28.7 (25.6)    & 55.6 (46.9)  \\
Phi-2                    & 2.7B      & -             & -         & 48.8 (45.1)    & 62.7 (52.9)  \\
\dscoderinst             & 1.3B      & {\markclosed} & 2B        & 65.2 (59.8)    & 63.9 (53.1)  \\ \midrule
\baselineds                & 1.3B             &  Evol-Instruct             &        0.3B             & 61.6 (57.3)    & 59.6 (49.1)  \\
\ewads                     & 1.3B             & Evol-Instruct            &        0.3B              &   \textbf{67.1} (63.4)  &   58.9 (48.4)  \\ \midrule
\oursmoe                   & 8$\times$1.3B    & Evol-Instruct           &        0.3B              & 65.2 (62.2)    & 60.4 (50.1)  \\
\oursmerge                 & 1.3B             & Evol-Instruct           &        0.3B               & \textbf{67.1} (\textbf{64.6})    & \textbf{60.4} (\textbf{50.1})  \\ \bottomrule
\end{tabular}
\caption{\label{tab:python-text2code}
\Passat{1} results of different code \llm{s} on \humaneval{}~(+) and \mbpp{}~(+) computed with greedy decoding, following the setting of prior works~\cite{wei2023magicoder, evalplus}. We report the results consistently from the \evalplus~\cite{evalplus} Leaderboard. Note that numbers in bold refer to the highest scores among all 1.3B models fine-tuned on public datasets, which is the same for all the other tables.
}
\end{table*}

\subsection{Experimental Setup}
\textbf{Training.} 
\dscoderbase 1.3B~\cite{guo2024deepseekcoder} is used as our main base code \llm. 
\evolcode, an open-source \evolinstruct~\cite{luo2023wizardcoder} dataset containing 110K samples, is used as our code instruction dataset.
\textbf{\oursmoe}, our \moe model upcycled from the base model, is implemented following Llama-MoE~\cite{llama-moe-2023}. 
It is constructed with 8 experts in one \moe layer and the top 6 experts\footnote{
6 is the best-performing number of activated experts per our \humanevalp{} experiments using top $\{2,4,6\}$ experts.} are activated for each token, including one shared expert. 
As such, we denote the model size of \textbf{\oursmoe} as 8$\times$1.3B.
Other training settings are detailed in Appendix \ref{sec:hyperparameter}. 
We further obtain \textbf{\oursmerge} by using the learned mixing coefficients to compile \moe layers inside \oursmoe to normal FFN layers. 
Note that \textbf{\oursmerge} is the final instruction-tuned code \llm we output, while \textbf{\oursmoe} is only an intermediate product of our \ours framework.

\textbf{Baselines.} 
To study the effectiveness of \ours, we build a baseline model, namely \textbf{\baselineds}, by directly performing SFT for \dscoderbase 1.3B on \evolcode. 
To compare \ours with \ewa~\cite{huang2023experts}, we also implement a baseline \textbf{\ewads} and instruction-tune it using the same hyperparameter setting as \baselineds, which is described in Appendix \ref{sec:hyperparameter}. More implementation details of \ewads are described in Appendix \ref{sec:ewa_details}. 
Furthermore, we incorporate multiple tiny open-source \llm{s} (<3B) as our baselines, including \dscoderbase 1.3B, \dscoderinst 1.3B~\cite{guo2024deepseekcoder}, Phi-2 2.7B, and \stablecoder 3B~\cite{stable-code-3b}.

\begin{table*}[h]
\centering
\begin{tabular}{@{}lcrrrrrrr@{}}
\toprule
\multirow{2}{*}{Model} & \multirow{2}{*}{Size} & \multicolumn{6}{c}{Programming Language}                                                      & \multirow{2}{*}{\textbf{Average}} \\ \cmidrule(lr){3-8}
                       &                       & C++           & PHP           & Java          & \multicolumn{1}{c}{JS}    & Swift         & Rust          &                          \\ \midrule
\dscoderbase    & 1.3B                  & 28.1          & 22.9         & 27.2          & 28.7           & 10.9          & 18.0          & 22.6                     \\ \midrule
\baselineds                 & 1.3B                  & 40.4          & 38.5          & \textbf{40.2} & 46.2          & 16.4          & 27.7          & 34.9                     \\
\ewads                    & 1.3B                  &     39.4      &     38.4      &      37.3     &      45.2     &     20.9      &   28.6   &    35.0                 \\ \midrule
\oursmoe                  & 8$\times$1.3B                & 42.2          & 42.2 & 35.4          & 49.8          & 24.7          & 30.6          & 37.5                     \\
\oursmerge                  & 1.3B                  & \textbf{42.7} & \textbf{41.5} & 36.0          & \textbf{49.7} & \textbf{25.3} & \textbf{32.1}          & \textbf{37.9}            \\ \bottomrule
\end{tabular}
\caption{\label{tab:multilang}
\Passat{1} results on \multiple~\cite{cassano2022multiple} following the same hyperparameter settings as prior works~\cite{wei2023magicoder, luo2023wizardcoder}: $\temperature=0.2$, $\topp=0.95$, $\maxLen=512$, and $\nsamples=50$. All models are evaluated using   \bigcodeharness{}~\cite{bigcode-evaluation-harness}.
}
\end{table*}

\begin{table*}[h]
\centering
\begin{tabular}{@{}lcrrrrrrrr@{}}
\toprule
\multirow{2}{*}{Model} & \multirow{2}{*}{Size} & \multicolumn{7}{c}{Data Science Library}                                                                      & \multirow{2}{*}{\textbf{Overall}} \\ \cmidrule(lr){3-9}
                       &                       &  \multicolumn{1}{c}{np}         &  \multicolumn{1}{c}{pd}        &  \multicolumn{1}{c}{plt}    &  \multicolumn{1}{c}{py}       &  \multicolumn{1}{c}{scp}         &  \multicolumn{1}{c}{tf}    &  \multicolumn{1}{c}{sk}       &                          \\ \midrule
\dscoderbase    & 1.3B                  &       25.1        &       5.8        &        34.5       &       12.7        &      9.8         &       11.1        &       12.7        &        16.4                  \\ \midrule
\baselineds                 & 1.3B                  & 30.9          & 17.0          & 40.5 & 32.7          & 18.3          & 21.1 & 24.4          & 25.9                     \\
\ewads                    & 1.3B                  &   32.9   &    19.4         &   \textbf{41.8}  &       25.7       &      17.7      &       \textbf{22.2}        &       33.0        &    27.8                     \\ \midrule
\oursmoe                  & 8$\times$1.3B                & 33.2          & 21.3          & 38.4          & 41.8          & 21.8          & 23.5    & 37.5          & 30.0               \\
\oursmerge                  & 1.3B                  & \textbf{32.9} & \textbf{20.2} & 38.9          & \textbf{41.4} & \textbf{21.1} & 16.9          & \textbf{37.5} & \textbf{29.3}            \\ \bottomrule
\end{tabular}
\caption{\label{tab:ds1000}
\Passat{1} results on \dsonek{} (completion format) with $\temperature=0.2$, $\topp=0.5$, $\maxLen=1024$, and $\nsamples=40$, following the same hyperparameter setting used in prior works~\cite{wei2023magicoder}.
}
\end{table*}

\subsection{Python Text-to-Code Generation}

\humaneval~\cite{chen2021evaluating} and \mbpp~\cite{austin2021program} benchmarks are the two most widely-used collections of Python code generation tasks. 
We further employ \humanevalp and \mbppp, which use more tests automatically generated by \evalplus~\cite{evalplus} for more rigorous evaluation. 
We leave the detailed description of \humaneval{(+)} and \mbpp{(+)} in Appendix \ref{sec:benchmarks}.

Table~\ref{tab:python-text2code} shows the \passat{1} results of different \llm{s}.
\ours achieves 67.1 \passat{1} on \humaneval and 64.6 \passat{1} on \humanevalp, which makes it the new state-of-the-art tiny code \llm{} (<3B). 
We can also observe that \oursmerge has a clear improvement over the \baselineds on both benchmarks, with 13\% and 2\% improvement on \humanevalp and \mbppp respectively. In contrast, \ewads not only underperforms \oursmerge on both benchmarks, but also fails to improve \baselineds on \mbpp{(+)}.
Surprisingly, \oursmerge even surpasses \oursmoe on \humaneval and \humanevalp, despite only using around $\sfrac{1}{8}\times$ parameters and around $\sfrac{1}{6}\times$ computations, which showcases the effectiveness of our simple learnable merging technique. More comprehensive experiments in Appendix~\ref{sec:statistic} demonstrate the statistical significance of the improvements brought by \ours. Furthermore, while \ours will inevitably introduce training overhead, our experiment in Appendix~\ref{sec:training_overhead} shows that \ours still significantly outperforms SFT using the same training budget, demonstrating the ability of \ours to unlock the power of code instruction tuning.

\subsection{Multilingual Code Generation}\label{sec:multiple}

We use \multiple~\cite{cassano2022multiple}, a multi-programming benchmark that supports 18 programming languages in addition to Python, to evaluate the multilingual ability and generalizability of \ours{}. 
Following previous work~\cite{wei2023magicoder}, we choose six representative programming languages in our evaluation for their distinct language features: Java, JavaScript, C++, PHP, Swift, and Rust.
Table \ref{tab:multilang} shows that, among all 1.3B models, \oursmerge achieves the best average multilingual performance and performs the best on five (out of six) programming languages and 
overall largely improves \baselineds{} which uses standard SFT.
Notably, the overall performance of \ewads is on par with \baselineds, indicating that \ewads{} fails to improve SFT on multilingual coding.
Appendix~\ref{sec:expert_analysis} further studies whether each expert in \oursmoe specializes differently in different programming languages.

\subsection{Code Generation for Data Science}

The \dsonek dataset~\cite{lai2022ds1000} is a collection of 1000 realistic data science coding problems ranging from seven popular data science libraries in Python, including Matplotlib (plt), NumPy (np), Pandas (pd), SciPy (scp), Scikit-Learn (sk), PyTorch (py), and TensorFlow (tf). 
We evaluate \ours on \dsonek{} to understand its effectiveness for practical data science engineering. 
We follow the evaluation setting of prior works~\cite{guo2024deepseekcoder, wei2023magicoder}. 
Table \ref{tab:ds1000} shows that \oursmerge achieves the best overall performance among all the evaluated 1.3B models. 
Specifically, \oursmerge consistently surpasses \baselineds among all the seven studied libraries and outperforms \ewads in general.

\section{Ablation Study}

\subsection{Effect of Shared Expert with Routing Weight Normalization}
\begin{table}[t]
\centering
\begin{tabular}{@{}lrr@{}}
\toprule
Model                                                                     & \multicolumn{1}{l}{HumanEval} & \multicolumn{1}{l}{HumanEval+} \\ \midrule
\baselineds                                                                     & 61.6                 & 57.3                  \\
\oursmoe                                                                     & \textbf{65.2}                 & \textbf{62.2}                  \\ \midrule
\begin{tabular}[c]{@{}l@{}}\oursmoe \\ \ \ - \scalebox{0.8}{Normalization}\end{tabular}                                                           & 63.4                          & 59.1                           \\ \midrule
\begin{tabular}[c]{@{}l@{}}\oursmoe \\ \ \ - \scalebox{0.8}{Shared Expert}\end{tabular} & 61.6                          & 56.7                           \\ \bottomrule
\end{tabular}
\caption{\label{tab:ablation-moe}
Ablation over the design of \oursmoe. "- Normalization" removes the routing weight normalization from the router, making it the same design as \mocle~\cite{gou2024mixture}. "- Shared Expert" removes the shared expert setting, making \oursmoe the same architecture as original \sparseupcycle~\cite{komatsuzaki2023sparse}.
}
\end{table}

We demonstrate the importance of the shared expert of \ours by comparing its performance with
the original \sparseupcycle~\cite{komatsuzaki2023sparse} baseline that does not employ any shared expert. As shown in Table \ref{tab:ablation-moe}, the performance of the original \sparseupcycle (with the "- Shared Expert" label) drops greatly compared with \oursmoe. Notably, the \sparseupcycle model performs even worse than \baselineds on \humanevalp, indicating its ineffectiveness for instruction tuning.

While the shared expert setting is also employed in most recent works~\cite{dai2024deepseekmoe, gou2024mixture}, their routing strategy will cause performance degradation due to the scale mismatch problem, which is handled by the routing weight normalization design in \ours. To demonstrate its importance, we conduct an ablation experiment by excluding it from \ours.
Table \ref{tab:ablation-moe} shows that, after removing routing weight normalization, the performance substantially decreases, despite still performing better than the original \sparseupcycle that does not use the shared expert setting.

\subsection{Effect of Merging Strategy}\label{sec:ablation_merging}
\begin{table}[]
\centering
\begin{tabular}{@{}lrr@{}}
\toprule
Model                                                         & \multicolumn{1}{c}{HumanEval} & \multicolumn{1}{c}{HumanEval+} \\ \midrule
\oursmoe                                                         & 65.2                          & 62.2                           \\
\oursmerge (INIT)                                                        & 66.5                 & 64.0                  \\
\oursmerge                                                         & \textbf{67.1}                 & \textbf{64.6}                  \\ \midrule
\begin{tabular}[c]{@{}l@{}}\oursmerge\\ \ \ - \scalebox{0.8}{Shared Expert Rate} \end{tabular} & 66.5                          & 64.0                           \\ \bottomrule
\end{tabular}
\caption{\label{tab:ablation-merge}
Ablation over the design of \oursmerge. "(INIT)" refers to directly using the initialized mixing coefficients to merge experts without training. "- Shared Rate" removes the shared rate setting from \oursmerge, which is the same as the learned soup~\cite{wortsman2022model}.
}
\end{table}

In this section, we demonstrate the effectiveness of our learnable merging technique by comparing it with (1) directly merging experts with initialized mixing coefficients, and (2) the learnable merging technique without the shared expert rate setting, which is the same setting as the learned soup in \modelsoup~\cite{wortsman2022model} and is described in Eq. (\ref{formula:merging1}) and Eq. (\ref{formula:merging2}). 
Specifically, we initialize the learnable mixing coefficient of the shared expert as 0.75 and that of the other 7 normal experts as $\frac{1}{28}$ for a fair comparison. As shown in Table \ref{tab:ablation-merge}, trained mixing coefficients outperform the initialized mixing coefficients for merging. Furthermore, removing the shared rate setting will degrade the performance of \oursmerge on both \humaneval and \humanevalp, demonstrating its importance. An ablation study on the shared expert rate in Appendix~\ref{sec:shared_expert_rate} further shows that (1) \oursmerge consistently outperforms \baselineds regardless of their shared expert rate, and (2) both the general knowledge learned in the shared expert and the specific knowledge learned in other experts are important and integral for better performance after merging.

\subsection{Effect of Code \llm Choice}
To show that the effectiveness of \ours is not dependent on any specific code \llm{s}, we apply \ours to \stablecoder 3B~\cite{stable-code-3b}, whose architecture is different from \dscoderbase 1.3B~\cite{guo2024deepseekcoder}, to study whether \ours can still improve the performance of this new model. The training settings are detailed in Appendix \ref{sec:stable_setting}. As shown in Table \ref{tab:ablation-stable}, \stablemerge significantly improves \baselinestable by 10\% on \humaneval and 11\% on \humanevalp respectively. Furthermore, \stablemerge consistently boosts the performance of \stablemoe while only using $\sfrac{1}{4}\times$ parameters and $\sfrac{1}{2}\times$ inference computations. These results show that the effectiveness of \ours is generalizable across different code \llm{s}.

\begin{table}[t]
\centering
\begin{tabular}{@{}lrr@{}}
\toprule
Model & \multicolumn{1}{c}{HumanEval} & \multicolumn{1}{c}{HumanEval+} \\ \midrule
\baselinestable   & 62.2                          & 56.1                           \\ \midrule
\stablemoe   & 64.0                            & 59.1                           \\
\stablemerge   &    \textbf{68.3}                           &    \textbf{62.2}                            \\ \bottomrule
\end{tabular}
\caption{\label{tab:ablation-stable}
Ablation over the effect of the base model by replacing \dscoderbase 1.3B with \stablecoder 3B. \ours can consistently improve the instruction tuning performance of different base code \llm{s}.
}
\end{table}

\section{Discussion}
\begin{table}[b]
\centering
\begin{tabular}{@{}lrr@{}}
\toprule
Model & \multicolumn{1}{c}{HumanEval} & \multicolumn{1}{c}{HumanEval+} \\ \midrule
\baselinedsscale   &      77.4                 &          70.7              \\ \midrule
\oursscalemoe   &         81.1                    &        75.6                    \\
\oursscalemerge   &    \textbf{81.7}                           &    \textbf{76.8}                            \\ \bottomrule
\end{tabular}
\caption{\label{tab:discussion-scaleup}
Experiments on scaling up \ours to 7B scale. It shows that \ours can also consistently improve the instruction tuning performance of 7B-level code \llm{s}.
}
\end{table}

\subsection{Scaling up \ours to 7B Scale}\label{sec:scale7b}
\begin{table*}[!t]
\centering
\begin{tabular}{@{}lccccc@{}}
\toprule
\multirow{2}{*}{Model} & \multicolumn{4}{c}{Discipline}                                                                                             & \multicolumn{1}{c}{\multirow{2}{*}{\textbf{Overall}}} \\ \cmidrule(lr){2-5}
                       & \multicolumn{1}{c}{Humanities} & \multicolumn{1}{c}{Social Science} & \multicolumn{1}{c}{STEM} & \multicolumn{1}{c}{Other} & \multicolumn{1}{c}{}    \\ \midrule
\baselinetinyllama & \textbf{25.38}                 & 23.30                               & 24.20                     & 26.78                     & 24.97                       \\ \midrule
\tinyllamamoe & 23.85                          & 26.32                              & 27.40                     & 28.03                     & 26.11                       \\
\tinyllamamerge & 23.91                          & \textbf{26.49}                     & \textbf{27.72}           & \textbf{28.29}            & \textbf{26.30}               \\ \bottomrule
\end{tabular}
\caption{\label{tab:discussion-generalizability}
Experiments on the generalizable effectiveness of \ours for general tasks in MMLU benchmark~\cite{hendrycks2021measuring}. It shows that \ours can improve the general instruction tuning performance of \llm{s}.
}
\end{table*}

We scale up \ours to 7B-level code \llm{s} by applying it to \dscoderbase 6.7B~\cite{guo2024deepseekcoder}. The training settings are detailed in Appendix~\ref{sec:dsscale_setting}. As shown in Table~\ref{tab:discussion-scaleup}, \oursscalemerge significantly improves \baselinedsscale by 6\% on \humaneval and 9\% on \humanevalp respectively. Moreover, \oursscalemerge further boosts the performance of \oursscalemoe with only $\sfrac{1}{8}\times$ parameters and $\sfrac{1}{2}\times$ computations during inference! These promising results demonstrate the consistent effectiveness of \ours on 7B-level code \llm{s}.

\subsection{Generalizability for General Tasks}

To demonstrate that \ours can also improve the performance of \llm{s} on general tasks across different domains, we apply \ours to general instruction tuning. We use \tinyllama 1.1B~\cite{zhang2024tinyllama} as the base model and use \evolgeneral~\cite{xu2023wizardlm} as the training dataset for general instruction tuning. Following existing work~\cite{zhang2024tinyllama}, we use MMLU~\cite{hendrycks2021measuring} with the 5-shot setting as our evaluation benchmark to evaluate the general performance of instruction-tuned \llm{s}. More training settings are detailed in Appendix \ref{sec:tinyllama_setting}. As shown in Table \ref{tab:discussion-generalizability}, \tinyllamamerge improves \baselinetinyllama by 5\% on MMLU in general, demonstrating the generalizable effectiveness of \ours for general instruction tuning.

\subsection{Preliminary Theoretical Explanation}
We provide a preliminary theoretical explanation of \ours by considering a simplified variant of it. Let’s start by analyzing the two major steps of \ours:
\begin{itemize}[leftmargin=1em]
    \setlength{\parskip}{2pt}
    \setlength\itemsep{0pt}
    \item \textbf{Step 1: Upcycling}. According to the scaling laws~\cite{kaplan2020scaling}, the upcycled \moe model performs better than the normal SFT dense model due to more trainable parameters.
    \item \textbf{Step 2: Merging}. We consider a simplified variant of \ours, where the upcycled \moe model (e.g., \oursmoe) can be viewed as the ensembling of two dense models and the merged dense model (e.g., \oursmerge) can be viewed as the merging of the same two dense models. More details are included in Appendix \ref{sec:theoratical_analysis}. As such, we can directly apply the theoretical analyzing process in Section 4 of Model Soups~\cite{wortsman2022model} to analyze the performance difference between the upcycled \moe model and the merged dense model, which is initially designed to analyze the performance difference between model ensembling and model merging. According to the analysis~\cite{wortsman2022model}, the convexity of the loss can help the merged dense model achieve a similar expected loss as that of the upcycled \moe model.
\end{itemize}

Overall, our preliminary theoretical explanation shows that (1) the \textbf{Upcycling} step improves the performance with more trainable parameters, and (2) the \textbf{Merging} step can provably maintain the performance of the aforementioned simplified \moe model with only dense-model compute.

\section{Conclusion}

This paper introduces \ours to unlock the power of code instruction tuning by simply merging upcycled \moe.
Similar to SFT, \ours starts with a dense \llm{} and outputs a fine-tuned dense \llm{} with the exact size and model structure.
Yet, \ours improves SFT by upcycling the pre-trained dense \llm{} to an \moe{} model for fine-tuning, after which we compile the \moe{} model back to an efficient dense \llm{} with a learnable merging mechanism.
As such, we unleash the performance limit of instruction tuning without any additional inference overhead.
Using the same training dataset, \ours outperforms SFT on a variety of benchmarks, including HumanEval(+), MBPP(+), \multiple{}, and \dsonek{}, from 2\% to 13\%.
By applying \ours to \dscoderbase{ 1.3B}, we create the new state-of-the-art tiny code \llm{} (<3B).
The final dense \llm{} produced by \ours preserves or even outperforms the full upcycled \moe which uses $8\times$ parameters as much as our final dense \llm{}.
\ours is fully orthogonal to the existing instruction tuners such as \evolinstruct{} and \ossinstruct{}, opening a new dimension for code instruction tuning.

\section*{Limitations}
To balance the general knowledge in the shared expert and the specific knowledge in other normal experts, we introduce a hyperparameter $\lambda$ in the merging process of \ours, which might slightly increase the efforts for hyperparameter search. It would be interesting to explore other hyperparameter-free techniques to tackle this challenge. Furthermore, while we have provided a preliminary theoretical explanation for the strong empirical performance of \ours, it would be interesting to provide a complete theoretical explanation in the future.

\section*{Acknowledgement}
We extend our special thanks to Terry Yue Zhuo for his assistance with the scale-up experiments on \dscoderbase 6.7B (\Cref{sec:scale7b}) after our submission. 
His contributions are good enough to merit authorship; however, due to the policy of ACL 2024, post-submission authorship changes are not permitted. 
As a result, we have included him in the author list of our arXiv version.
We also thank Sea AI Lab and Dr. Qian Liu for their valuable feedback and computing resource assistance.
We appreciate all the reviewers for their insightful comments. 
This work was partially supported by NSF grant CCF-2131943, as well as Kwai Inc.

\bibliography{custom}

\appendix

\section{Appendix for "\ours: Unlocking the Power of Code Instruction Tuning by Simply Merging Upcycled Mixture-of-Experts"}
\label{sec:appendix}

\subsection{Training Settings for \dscoderbase 1.3B}\label{sec:hyperparameter}
We use a batch size of 64 and a learning rate of 5e-5 with a linear scheduler to fine-tune \textbf{\oursmoe} for 4 epochs with 500 warmup steps, following the implementation of previous work~\cite{wei2023magicoder}. We further use a batch size of 64, a shared expert rate $\lambda$ of 0.75, and a learning rate of 1e-5 with a linear schedule to fine-tune the learnable mixing coefficients for experts in the instruction-tuned \textbf{\oursmoe} on the same instruction-tuning dataset for 1 epoch with 125 warmup steps. Detailedly, we use Softmax to keep the sum of the mixing coefficients of the other 7 normal experts as 0.25. For \textbf{\baselineds} and \textbf{\ewads}, we use the same hyperparameter setting as \ours, where the batch size is 64 and the learning rate is 5e-5 with a linear scheduler. Because \ours is trained for 4 epochs during upcycling and 1 epoch during merging, for a fair comparison, we train \baselineds and \ewads for 5 (= 4 + 1) epochs with 625 warmup steps.

\subsection{Implementation details of \ewa}\label{sec:ewa_details}
\begin{figure}[ht]
\centering
\includegraphics[width=1.0\linewidth]{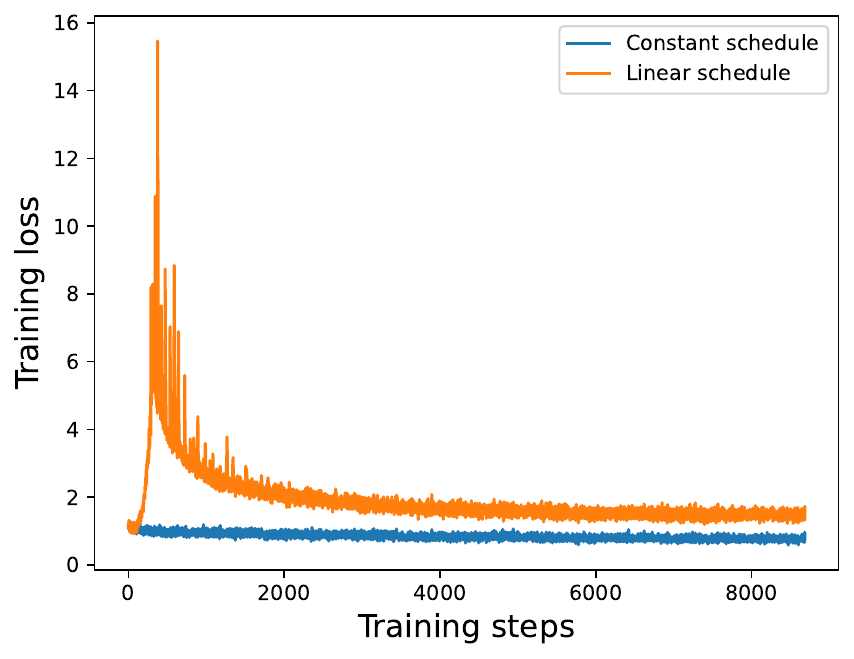}
\caption{Training loss curve of \ewa with the constant schedule setting and the linear schedule setting.}
\label{fig:loss_ewa}
\end{figure}

Because \ewa~\cite{huang2023experts} does not release their implementation, we implement \ewa by ourselves, including the constant schedule setting and the linear schedule setting. We use a share rate $\beta$ of 0.3, following the original setting of \ewa. While \ewa with the constant schedule setting achieves reasonable performance in our evaluation, the training loss of \ewa with the linear schedule setting is unstable during instruction tuning, as shown in Figure \ref{fig:loss_ewa}, and thus cannot achieve reasonable performance. As a result, we report the results of \ewa with the constant schedule setting in Section \ref{sec:experiment}.

\subsection{Details of \humaneval{(+)} and \mbpp{(+)}}\label{sec:benchmarks}
In these benchmarks, each task consists of a task description in English, which is sent to \llm{s} as the prompt, and \llm{s} are expected to generate the corresponding code to satisfy the requirements in the description. While these benchmarks provide a handful of test cases to validate the correctness of the generated code, these tests are often insufficient for more rigorous evaluation. As such, \humanevalp and \mbppp proposed by \evalplus~\cite{evalplus} are usually used to evaluate the correctness of the generated code, which provides 80×/35× more tests compared with the original benchmarks.

\subsection{Statistical Significance Analysis}\label{sec:statistic}
\begin{table}[t]
\centering
\begin{tabular}{@{}lrr@{}}
\toprule
Model                         & \multicolumn{1}{c}{HumanEval} & \multicolumn{1}{c}{HumanEval+} \\ \midrule
\baselineds                      & 61.6                          & 57.2                           \\
\ewads & 62.7                          & 58.8                           \\
\oursmerge                      & \textbf{64.5}                 & \textbf{60.9}                  \\ \bottomrule
\end{tabular}
\caption{\label{tab:discussion-stat-mean}
Average pass@1 results of 200 experiments on \humaneval{}~(+) computed with sampling. \ours clearly outperforms both \ewads and \baselineds.
}
\end{table}

\begin{table}[t]
\centering
\begin{tabular}{@{}lrr@{}}
\toprule
Model            & \multicolumn{1}{c}{HumanEval} & \multicolumn{1}{c}{HumanEval+} \\ \midrule
\scalebox{0.9}{\oursmerge vs. \ewads} & 2.6e-18                       & 8.0e-23                        \\
\scalebox{0.9}{\oursmerge vs. \baselineds} & 9.6e-30                       & 3.7e-33                        \\ \bottomrule
\end{tabular}
\caption{\label{tab:discussion-stat-p}
$p$-values for \oursmerge vs. \ewads and \oursmerge vs. \baselineds in 200 experiments on \humaneval{}~(+) conducted using sampling. Results show that improvements brought by \ours are statistically significant.
}
\end{table}

In our main experiments, we follow prior works~\cite{wei2023magicoder, lozhkov2024starcoder} to conduct experiments on \humaneval{(+)} using greedy decoding. To demonstrate the statistical significance of our improvements, we change our setting from greedy decoding to sampling. In detail, to conduct one experiment on \humaneval{(+)}, the model will sample one solution for each problem in \humaneval{(+)} with top $p$ = 0.95 and temperature = 0.8, which is the same setting used in prior works~\cite{evalplus, chen2021evaluating}.

Following prior work~\cite{evalplus}, we repeat this experiment 200 times for three techniques: \oursmerge, \ewads, and \baselineds. \ewads is included because it is the best-performing baseline in our main experiment. We first compute their average pass@1 performance in these 200 experiments. As is shown in Table \ref{tab:discussion-stat-mean}, \oursmerge outperforms both \ewads and \baselineds.

Furthermore, we use the Wilcoxon signed-rank test~\cite{Wilcoxon1945IndividualCB, dror-etal-2018-hitchhikers}, a widely used statistical test, to check if the improvements brought by \ours are statistically significant. As shown in Table \ref{tab:discussion-stat-p}, the $p$-values for both \oursmerge vs. \ewads and \oursmerge vs. \baselineds are much smaller than both 0.0025 (the significance level recommended for NLP work~\cite{sogaard-etal-2014-whats}) and 0.05 (the most common significance level), demonstrating the statistical significance of the improvements brought by \ours.

\subsection{Training Overhead Analysis}\label{sec:training_overhead}
\begin{table}[t]
\centering
\begin{tabular}{@{}lrr@{}}
\toprule
Model                           & \multicolumn{1}{c}{HumanEval} & \multicolumn{1}{c}{HumanEval+} \\ \midrule
\begin{tabular}[c]{@{}l@{}}\baselineds\\ \ \ \scalebox{0.9}{w/ same steps} \end{tabular}  & 61.6                          & 57.3                           \\
\begin{tabular}[c]{@{}l@{}}\baselineds\\ \ \ \scalebox{0.9}{w/ same budget} \end{tabular} & 62.2                          & 57.3                           \\ \midrule
\oursmerge                        & \textbf{67.1}                 & \textbf{64.6}                  \\ \bottomrule
\end{tabular}
\caption{\label{tab:discussion-overhead}
Experiments on the effect of training overhead. For our two SFT baselines, "w/ same steps" refers to one SFT baseline using the same training steps as \ours while "w/ same budget" refers to the other SFT baseline using the same training budget as \ours. \ours can consistently outperform both SFT baselines to a large extent, further demonstrating the ability of \ours to unlock the power of code instruction tuning.
}
\end{table}

Compared with SFT, \ours will inevitably introduce additional overhead in the training process because \ours needs to fine-tune the upcycled \moe model, which contains more parameters than the original dense model and thus requires more computation. In contrast, the normal SFT technique only needs to fine-tune the original dense model. To better understand the effect of such overhead, we conduct an experiment using the same training budget (i.e., the same GPU hours) instead of the same training steps for the normal SFT baseline. As shown in Table \ref{tab:discussion-overhead}, although sharing the same training budget as \oursmerge, the performance of \baselineds is still significantly worse than that of \oursmerge, demonstrating the ability of \ours to unlock the power of code instruction tuning using the same training budget.

\subsection{Expert Specialization Analysis}\label{sec:expert_analysis}
\begin{figure*}[ht]
\centering
\includegraphics[width=1.0\linewidth]{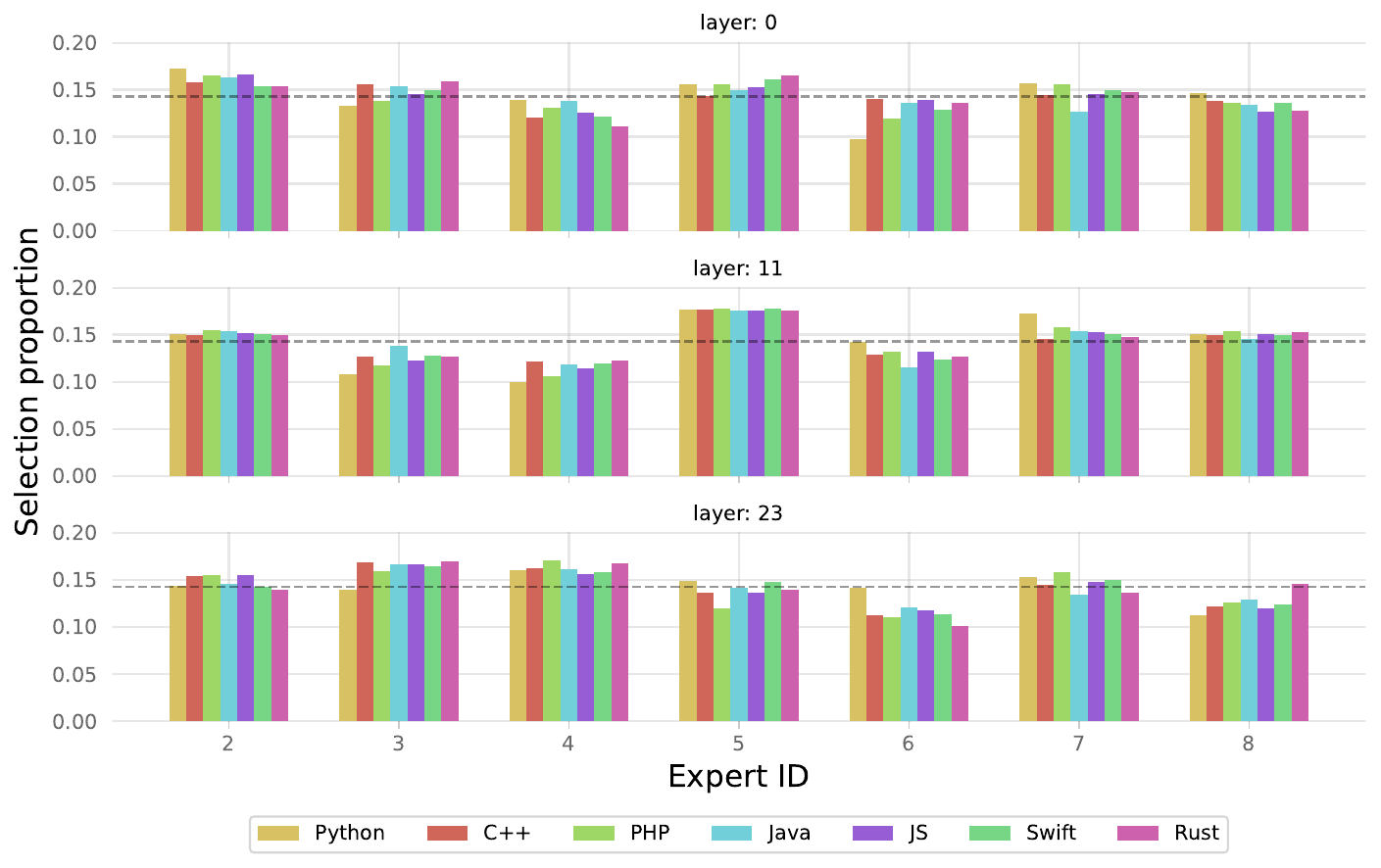}
\caption{Proportion of tokens assigned to each expert on different programming languages from \multiple (including Python) for layers 0, 11, and 23. The shared expert $\mbox{FFN}_1$ is excluded from the chart because all the tokens are always assigned to it. The gray vertical line $\frac{1}{7}$ is the proportion expected with the uniform sampling.}
\label{fig:analysis}
\end{figure*}

Inspired by recent works~\cite{jiang2024mixtral, xue2024openmoe}, we analyze whether each expert in \oursmoe has different specializations in different programming languages by visualizing the routing decision of the tokens from different programming languages in the \multiple benchmark (including Python). We collect the routing decision for the \multiple benchmark when conducting experiments in Section \ref{sec:multiple}. For Python, we collect the routing decision by running \humaneval experiment following the same setting used in Section \ref{sec:multiple}. Following the analysis setting of recent work~\cite{jiang2024mixtral}, we get the visualization results from layers 0, 11, and 23 in \oursmoe, where layer 0 and layer 23 are the first and the last layers of \oursmoe. As shown in Figure \ref{fig:analysis}, we do not observe any obvious
patterns in the assignment of experts based on the type of programming languages, which is in line with the findings reported by recent works~\cite{jiang2024mixtral, xue2024openmoe}. 

\subsection{Effect of Shared Expert Rate}\label{sec:shared_expert_rate}

We further study the effect of the shared expert rate $\lambda$ on the performance of the final merged dense model. We evenly choose five shared expert rates, including 0.00, 0.25, 0.50, 0.75, and 1.00, to perform the learnable merging process and evaluate each merged dense model accordingly. Note that 0.75 is the default shared expert rate used in our main experiments. If the shared expert rate is 0.00, it means that the shared expert is ignored when constructing the merged dense model from the upcycled \moe model; if the shared expert rate is 1.00, it means that the final dense model is built by simply extracting the shared expert from the upcycled \moe model. As shown in Table \ref{tab:ablation-shared-expert-rate}, there are mainly three interesting observations:
\begin{itemize}[leftmargin=1em]
    \setlength{\parskip}{2pt}
    \setlength\itemsep{0pt}
    \item The performance of the final merged dense model improves gradually when the shared expert rate grows from 0.00 to 0.75, indicating that \textbf{general knowledge learned by the shared expert is important for better performance}.
    \item The performance of the final merged dense model drops significantly when the shared expert rate grows from 0.75 to 1.00, showing that \textbf{specific knowledge learned by other experts is also integral} and ignoring them will lead to a significant performance drop.
    \item All the final merged dense models \textbf{consistently outperform the normal SFT baseline regardless of their shared expert rate}, further demonstrating the effectiveness of \ours.
\end{itemize}

\begin{table}[t]
\centering
\begin{tabular}{@{}lcrr@{}}
\toprule
Model                       & \multicolumn{1}{c}{$\lambda$} & \multicolumn{1}{c}{HumanEval} & \multicolumn{1}{c}{HumenEval+} \\ \midrule
\baselineds                 & -                             & 61.6                          & 57.3                           \\ \midrule
\multirow{5}{*}{\oursmerge} & 0.00                          & 62.8                          & 59.8                           \\
                                           & 0.25                          & 64.6                          & 61.0                           \\
                                           & 0.50                          & 65.9                          & 62.8                           \\
                                           & \textbf{0.75}                          & \textbf{67.1}                 & \textbf{64.6}                  \\
                                           & 1.00                          & 63.4                          & 60.4                           \\ \bottomrule
\end{tabular}
\caption{\label{tab:ablation-shared-expert-rate}
Ablation over the effect of the shared expert rate $\lambda$ in our learnable merging technique. \ours can consistently outperform the normal SFT baseline regardless of the shared expert rate, while $\lambda=0.75$ is the optimal setting in our experiments.
}
\end{table}

\subsection{Training Settings for \stablecoder 3B}\label{sec:stable_setting}
We use \evolcode as the training dataset. Since \stablecoder 3B is the base model, we upcycle a new \moe model from the base model, namely \textbf{\stablemoe}. We construct \stablemoe with 4 experts in one \moe layer, where the top 2 experts are activated for each token, including one shared expert. Consequently, the size of \stablemoe can be described as 4$\times$3B. We use a batch size of 64 and a learning rate of 5e-5 with a linear scheduler to fine-tune \stablemoe for 4 epochs with 500 warmup steps. Similar to \oursmerge, we obtain \textbf{\stablemerge} by learning mixing coefficients to merge \moe layers inside \stablemoe as normal FFN layers, which is fine-tuned with a batch size of 64, a shared expert rate $\lambda$ of 0.85, and a learning rate of 1e-5 with a linear schedule for 1 epoch with 125 warmup steps. Our baseline model, namely \textbf{\baselinestable}, is fine-tuned for 5 (= 4 + 1) epochs with a batch size of 64, a learning rate of 5e-5, and 625 warmup steps for a fair comparison.

\subsection{Training Settings for \dscoderbase 6.7B}\label{sec:dsscale_setting}
We use \evolcode as the training dataset. We upcycle a new \moe model from \dscoderbase 6.7B, namely \textbf{\oursscalemoe}. We construct \oursscalemoe with 8 experts in one \moe layer, where the top 2 experts are activated for each token, including one shared expert. As such, \oursscalemoe includes 8$\times$6.7B parameters. We use a batch size of 64 and a linear scheduler to fine-tune \oursscalemoe for 4 epochs with 500 warmup steps. We choose the best-performing learning rate from $\{2e-5, 5e-5\}$ for \oursscalemoe. Because the FFN weights of \oursscalemoe are too large to fit in our GPU memory, during our merging step, we realize that one part of computation in the training of \oursscalemerge has to be moved to CPUs, which significantly slows down the training speed. Consequently, we use a batch size of 16, a shared expert rate $\lambda$ of 0.75, a constant learning rate of 1e-4, and 400 training steps in merging to obtain \oursscalemerge. Our baseline model, namely \textbf{\baselinedsscale}, is fine-tuned for 5 epochs with a batch size of 64 and 625 warmup steps for a fair comparison. We also choose the best-performing learning rate from $\{2e-5, 5e-5\}$ for \baselinedsscale.

\subsection{Training Settings for \tinyllama 1.1B}\label{sec:tinyllama_setting}
Using \tinyllama 1.1B as the base model, we upcycle a new \moe model, namely \textbf{\tinyllamamoe}, from the pre-trained dense model. Following the setting of \oursmoe, we construct \tinyllamamoe with 8 experts in one \moe layer, where the top 6 experts are activated for each token, including one shared expert. As such, the number of parameters for \tinyllamamoe can be written as 8$\times$1.1B. We use a batch size of 64 and a learning rate of 5e-5 with a linear scheduler to fine-tune \tinyllamamoe for 4 epochs with 240 warmup steps. To obtain \textbf{\tinyllamamerge}, we learn mixing coefficients to merge \moe layers inside \tinyllamamoe by fine-tuning them with a batch size of 64, a shared expert rate $\lambda$ of 0.85, and a learning rate of 2e-5 with a linear schedule for 1 epoch with 60 warmup steps. For a fair comparison, we fine-tune a baseline model \textbf{\baselinetinyllama} for 5 (= 4 + 1) epochs with a batch size of 64, a learning rate of 5e-5, and 300 warmup steps.

\subsection{Details of Preliminary Theoretical Explanation}\label{sec:theoratical_analysis}
We consider a simplified variant of \ours as follows:
\begin{itemize}[leftmargin=1em]
\setlength{\parskip}{2pt}
\setlength\itemsep{0pt}
\item The original dense model is a one-layer transformer model, which contains one attention layer connected with one feed-forward network (FFN) layer. As such, the upcycled \moe model is also a one-layer transformer model, containing one attention layer connected with an \moe layer.
\item The upcycled \moe model only has two experts ($\textbf{e}_1$ and $\textbf{e}_2$), both of which are always selected for processing the input tokens.
\item The router in the \moe model assigns constant weights to each expert, regardless of the input token. Consequently, the output of the \moe layer for the $t$-th token $\textbf{h}_t$ can be represented as $(1-\alpha) \textbf{e}_1(\textbf{u}_t) + \alpha \textbf{e}_2(\textbf{u}_t)$, where $1-\alpha$ is the router weight assigned to $\textbf{e}_1$, $\alpha$ is the router weight assigned to $\textbf{e}_2$, and $\textbf{u}_t$ is the input of the \moe layer for the $t$-th token.
\item We simplify the process of merging the \moe model back to a dense model as $\textbf{W}_{\textbf{e}_{\alpha}} = (1-\alpha) \textbf{W}_{\textbf{e}_1} + \alpha \textbf{W}_{\textbf{e}_2}$, where $\textbf{W}_\textbf{e}$ refers to the weight of $\textbf{e}$ and $\textbf{e}_{\alpha}$ refers to the weight of the FFN in the merged dense model.
\end{itemize}

In this simplified scenario, if we denote $f(x;\theta)$ as the output of the model $\theta$ for the input $x$, the output of this simplified MoE model for input token $x$ can be represented as $f(x;\theta_{\textbf{MoE}})$. Interestingly, if we define two new dense models $\theta_1$ and $\theta_2$, where $\theta_1$ and $\theta_2$ both use the attention layer of this \moe model as their attention layer while using $\textbf{e}_1$ and $\textbf{e}_2$ as their FFN layer respectively, $f(x;\theta_{\textbf{MoE}})$ can be represented as $(1-\alpha) f(x;\theta_1) + \alpha f(x;\theta_2)$. Consequently, the computation process of this simplified \moe model can be viewed as ensembling the outputs of two dense models $\theta_1$ and $\theta_2$. Meanwhile, the process of merging the upcycled \moe model back to a dense model in this simplified scenario can be represented as $\theta_{\alpha} = (1-\alpha) \theta_1 + \alpha \theta_2$, which can be viewed as the model merging of the same two dense models $\theta_1$ and $\theta_2$.

\end{document}